\def\year{2019}\relax
\documentclass[letterpaper]{article} 
\usepackage{aaai19}  
\usepackage{times}  
\usepackage{helvet}  
\usepackage{courier}  
\usepackage{url}  
\usepackage{graphicx}  
\usepackage{subcaption}

\usepackage{ amssymb }

\begin{document}
%
\title{The Goldilocks zone: Towards better understanding of neural network loss landscapes}
\author{Stanislav~Fort \\
	Physics Department / KIPAC\\
	Stanford University\\
	382 Via Pueblo, Stanford, CA 94305 \\
	\texttt{sfort1@stanford.edu} \\
	\And
	Adam Scherlis \\
	Physics Department / SITP \\
	Stanford University \\
	382 Via Pueblo, Stanford, CA 94305\\
	\texttt{scherlis@stanford.edu} \\}
\maketitle
\begin{abstract}
We explore the loss landscape of fully-connected and convolutional neural networks using random, low-dimensional hyperplanes and hyperspheres. Evaluating the Hessian, $H$, of the loss function on these hypersurfaces, we observe 1) an unusual excess of the number of positive eigenvalues of $H$, and 2) a large value of $\mathrm{Tr}(H) / ||H||$ at a well defined range of configuration space radii, corresponding to a thick, hollow, spherical shell we refer to as the \textit{Goldilocks zone}. We observe this effect for \mbox{fully-connected} neural networks over a range of network widths and depths on MNIST and \mbox{CIFAR-10} datasets with the $\mathrm{ReLU}$ and $\tanh$ non-linearities, and a similar effect for convolutional networks. Using our observations, we demonstrate a close connection between the Goldilocks zone, measures of local convexity/prevalence of positive curvature, and the suitability of a network initialization. We show that the high and stable accuracy reached when optimizing on random, low-dimensional hypersurfaces is directly related to the overlap between the hypersurface and the Goldilocks zone, and as a corollary demonstrate that the notion of intrinsic dimension is initialization-dependent. We note that common initialization techniques initialize neural networks in this particular region of unusually high convexity/prevalence of positive curvature, and offer a geometric intuition for their success. Furthermore, we demonstrate that initializing a neural network at a number of points and selecting for high measures of local convexity such as $\mathrm{Tr}(H) / ||H||$, number of positive eigenvalues of $H$, or low initial loss, leads to statistically significantly faster training on MNIST. Based on our observations, we hypothesize that the Goldilocks zone contains an unusually high density of suitable initialization configurations.  
\end{abstract}
\section{Introduction}
\subsection{Objective Landscape}
A neural networks is fully specified by its architecture -- connections between neurons -- and a particular choice of weights $\{ W \}$ and biases $\{ b \}$ -- free parameters of the model. Once a particular architecture is chosen, the set of all possible value assignments to these parameters forms the \textit{objective landscape} -- the configuration space of the problem. Given a specific dataset and a task, a loss function $L$ characterizes how unhappy we are with the solution provided by the neural network whose weights are populated by the parameter assignment $P$. Training a neural network corresponds to optimization over the objective landscape, searching for a point -- a configuration of weights and biases -- producing a loss as low as possible.

The dimensionality, $D$, of the objective landscape is typically very high, reaching hundreds of thousands even for the most simple of tasks. Due to the complicated mapping between the individual weight elements and the resulting loss, its analytic study proves challenging. Instead, the objective landscape has been explored numerically. The high dimensionality of the objective landscape brings about considerable geometrical simplifications that we utilize in this paper.

\begin{figure}[ht]
	\centering
		\includegraphics[width=0.8\linewidth]{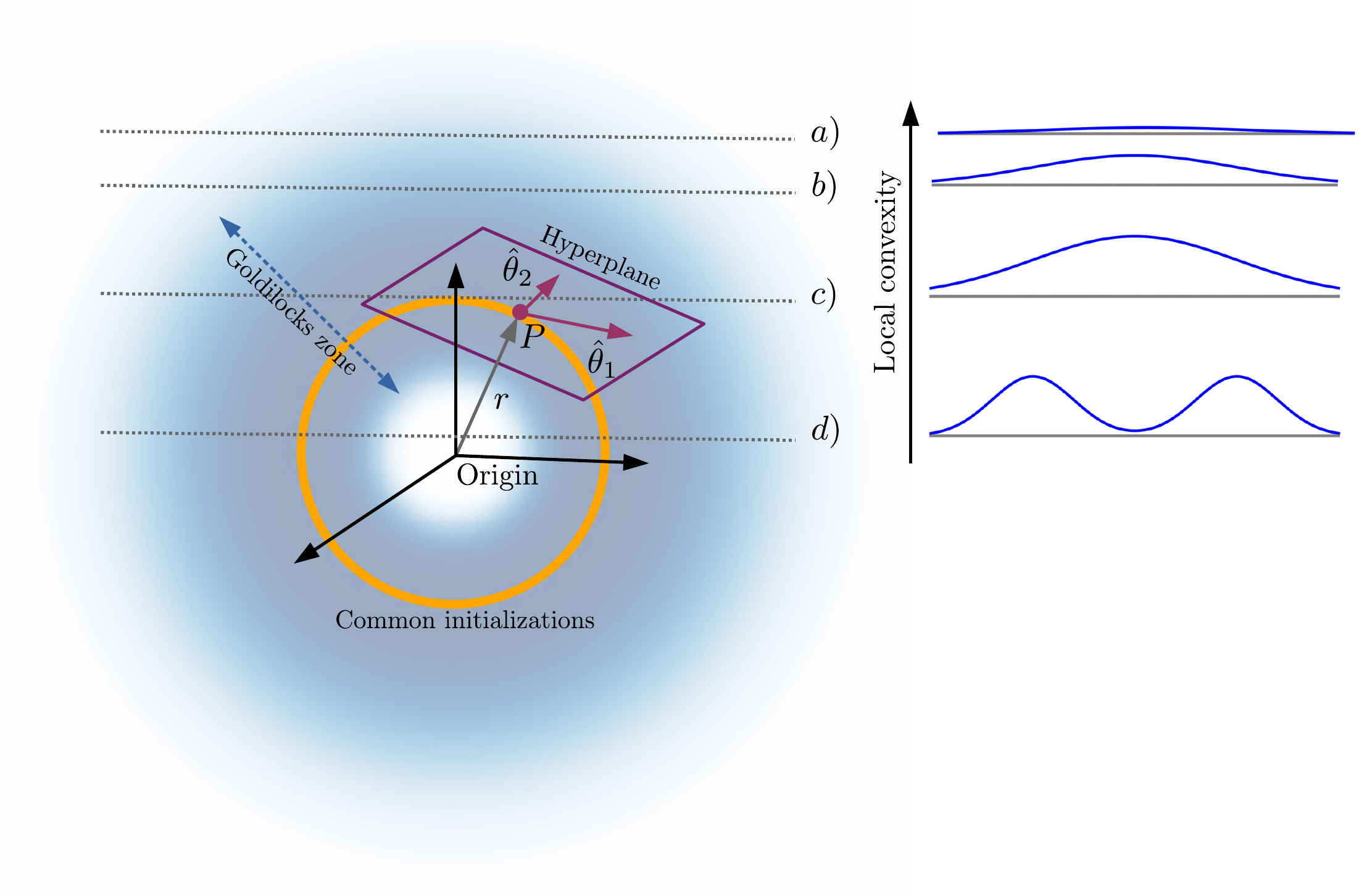}
	\caption{An illustration of the loss landscape. The thin spherical shell on which common initialization procedures initialize neural networks is shown in yellow, and its radius in gray. A random, low-dimensional hyperplane intersecting such a configuration $P$ is shown in purple, together with two of its coordinate directions $\hat{\theta}_1$ and $\hat{\theta}_2$. A thick shell -- the \textit{Goldilocks zone} -- of unusually high convexity/prevalence of positive curvature (e.g. unusual behavior of $\mathrm{Tr}(H)/||H||$) is shown as blue shading. Cuts through the region by hyperplanes at 4 different radii are shown schematically. The radii of common initializations lie well within the Goldilocks zone, therefore a random hyperplane perpendicular to $\hat{r}$ is bound to have a significant overlap with the zone.
		}
	\label{fig:sphere}
\end{figure}

\subsection{Related Work}
Neural network training is a large-scale non-convex optimization task, and as such provides space for potentially very complex optimization behavior. \cite{line_goodfellow}, however, demonstrated that the structure of the objective landscape might not be as complex as expected for a variety of models, including fully-connected neural networks \cite{fully_connected_original} and convolutional neural networks \cite{lecun2010convolutional}. They showed that the loss along the direct path from the initial to the final configuration typically decreases monotonically, encountering no significant obstacles along the way. The general structure of the objective landscape has been a subject of a large number of studies \cite{DBLP:journals/corr/ChoromanskaHMAL14,DBLP:journals/corr/KeskarMNST16}.

A vital part of successful neural network training is a suitable choice of initialization. Several approaches have been developed based on various assumptions, most notably the so-called \textit{Xavier initialization} \cite{Xavier}, and \textit{He initialization} \cite{He}, which are designed to prevent catastrophic shrinking or growth of signals in the network. We address these procedures, noticing their geometric similarity, and relate them to our theoretical model and empirical findings.

A striking recent result by \cite{intrinsic} demonstrates that we can restrict our degrees of freedom to a \textit{randomly} oriented, low-dimensional hyperplane in the full configuration space, and still reach almost as good an accuracy as when optimizing in the full space, provided that the dimension of the hyperplane $d$ is larger than a small, task-specific value $d_\mathrm{intrinsic} \ll D$, where $D$ is the dimension of the full space. We address and extend these observations, focusing primarily on the surprisingly low variance of accuracies on random hyperplanes of a fixed dimension.
\begin{figure}[ht]
	\centering
	\begin{subfigure}{.23\textwidth}
		\centering
		\includegraphics[width=1.0\linewidth]{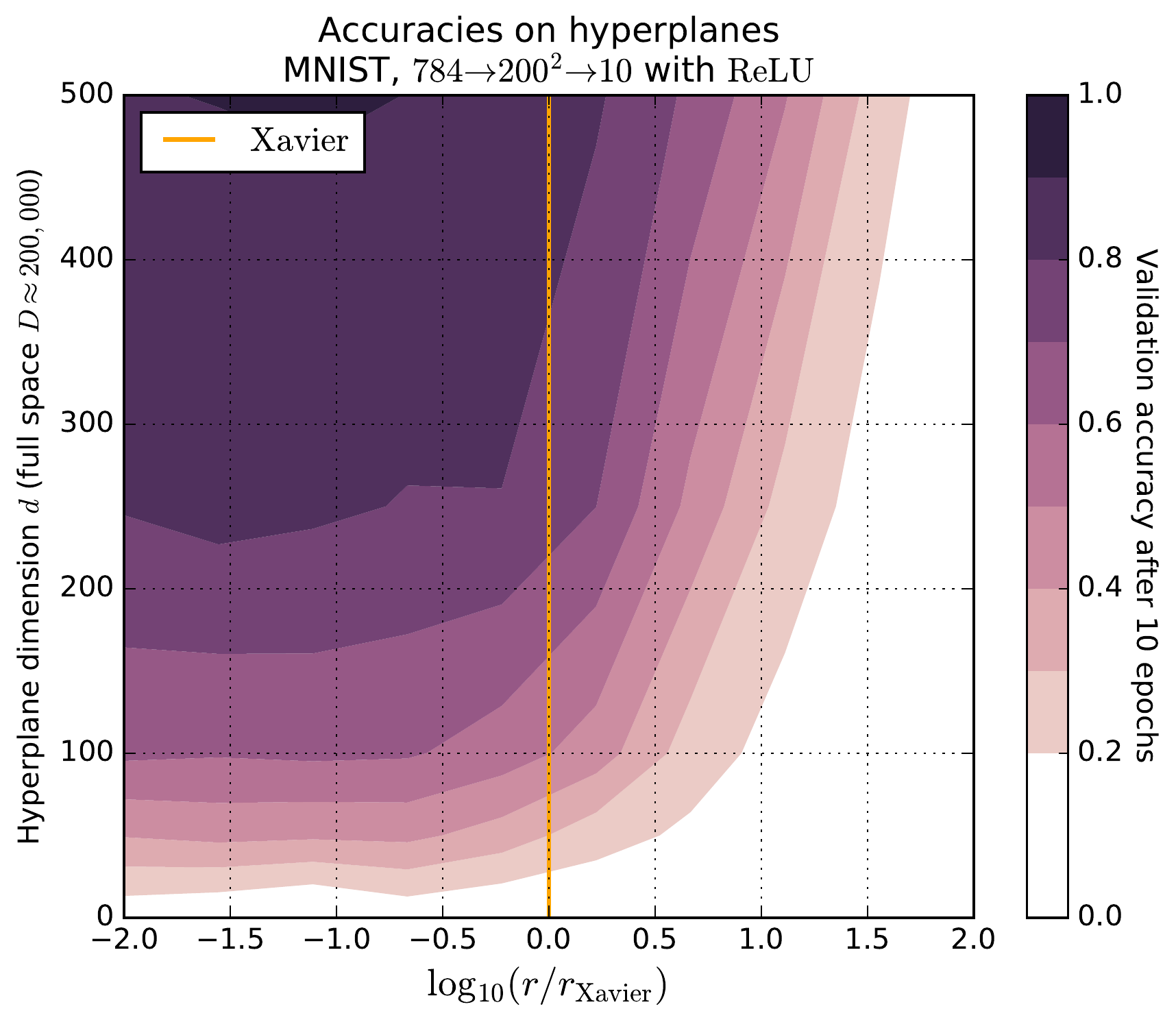}
		\caption{Accuracies reached on hyper\textit{planes} of different dimensions and distances from origin. The contours show the validation accuracy reached when optimizing on random hyperplanes initialized at distance $r$ from the origin. Consistently with our hypothesis, $r<r_{\mathrm{Xavier}}$ leads to an equivalent performance to $r=r_\mathrm{Xavier}$, as the hyperplanes intersect the Goldilocks zone. For $r>r_{\mathrm{Xavier}}$, the performance drops as hyperplanes no longer intersect the Goldilocks zone.}
		\label{fig:accs_contours}
	\end{subfigure}\hfill
	\begin{subfigure}{.23\textwidth}
		\centering
		\includegraphics[width=1.0\linewidth]{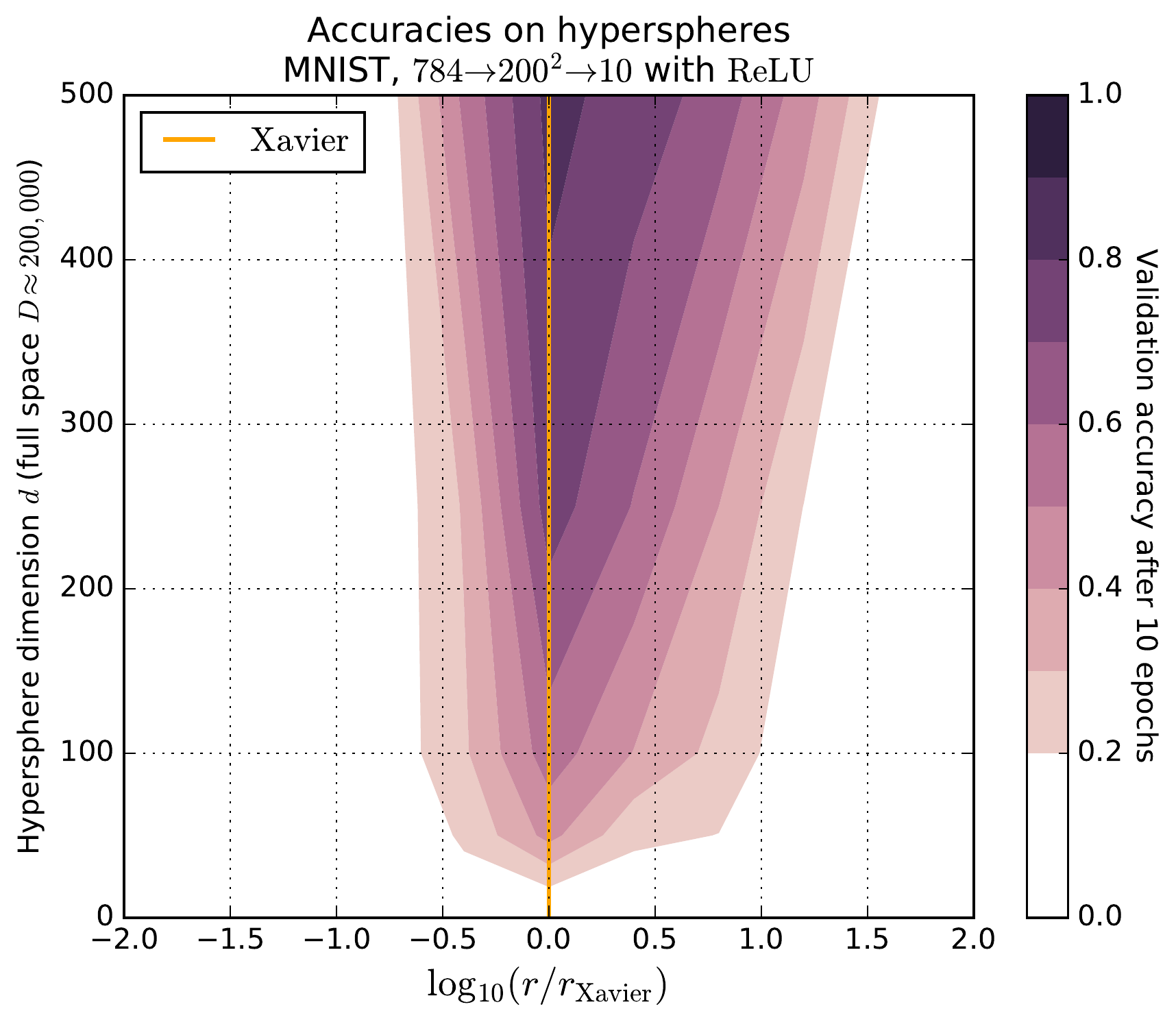}
		\caption{Accuracies reached on hyper\textit{spheres} of different dimensions and radii from origin. The contours show the validation accuracy reached when optimizing on random hyperspheres initialized at distance $r$ from the origin. The optimization was constrained to stay at this radius. Consistently with our hypothesis, $r<r_{\mathrm{Xavier}}$ as well as $r>r_{\mathrm{Xavier}}$ lead to poor performance, as the spherical surface on which optimization takes place does not intersect the Goldilocks zone.}
		\label{fig:accs_contours_spheres}
	\end{subfigure}
	\caption{Accuracies reached on random, low-dimensional hyper\textit{planes} (panel \ref{fig:accs_contours}) and hyper\textit{spheres} (panel \ref{fig:accs_contours_spheres}). The contours show the validation accuracy reached when optimizing on random hypersurfaces, confirming that good initial points are distributed on a thick, hollow, spherical shell, as illustrated in Figure~\ref{fig:sphere}. Plots of accuracy as a function of dimension presented in \cite{intrinsic} correspond to sections along the yellow vertical line and we therefore extend them. Consequently, this shows that the \textit{intrinsic dimension} \cite{intrinsic} of a problem is radius-dependent, and therefore initialization-dependent.}
	\label{fig:contours}
\end{figure}
\subsection{Our Contributions}
In this paper, we constrain our optimization to randomly chosen $d$-dimensional hyper\textit{planes} and hyper\textit{spheres} in the full $D$-dimensional configuration space to empirically explore the structure of the objective landscape of neural networks. Furthermore, we provide a step towards an analytic description of some of its general properties. Evaluating the Hessian, $H$, of the loss function on randomly oriented, low-dimensional hypersurfaces, we observe 1) an unusual excess of the number of positive eigenvalues of $H$, and 2) a large value of $\mathrm{Tr}(H) / ||H||$ at a well defined range of configuration space radii, corresponding to a thick, hollow, spherical shell of unusually high local convexity/prevalence of positive curvature we refer to as the \textit{Goldilocks zone}.

Using our observations, we demonstrate a close connection between the Goldilocks zone, measures of local convexity/prevalence of positive curvature, the suitability of a network initialization, and the ability to optimize while constrained to low-dimensional hypersurfaces.
Extending the experiments in \cite{intrinsic} to different radii and generalizing to hyper\textit{spheres}, we are able to demonstrate that the main predictor of the success of optimization on a $(d \ll D)$-dimensional sub-manifold is the amount of its overlap with the Goldilocks zone. We therefore demonstrate that the concept of \textit{intrinsic dimension} from \cite{intrinsic} is radius- and therefore initialization-dependent. Using the realization that common initialization techniques \cite{Xavier,He}, due to properties of high-dimensional Gaussian distributions, initialize neural networks in the same particular region, we conclude that the Goldilocks zone contains an exceptional amount of very suitable initialization configurations.

As a byproduct, we show hints that initializing a neural network at a number of points at a given radius, and selecting for high number of positive Hessian eigenvalues, high $\mathrm{Tr}(H) / ||H||$, or low initial validation loss (they are all strongly correlated in the Goldilocks zone), leads to statistically significantly faster convergence on MNIST. This further strengthens the connection between the measures of local convexity and the suitability of an initialization. 

Our empirical observations are consistent across a range of network depths and widths for fully-connected neural networks with the $\mathrm{ReLU}$ and $\tanh$ non-linearities on the MNIST and CIFAR-10 datasets. We observe a similar effect for CNNs. The wide range of scenarios in which we observe the effect suggests its generality.

This paper is structured as follows: We begin by introducing the notion of random hyperplanes and continue with building up the theoretical basis to explain our observations in Section~\ref{sec:th}. We report the results of our experiments and discuss their implications in Section~\ref{sec:results}. We conclude with a summary and future outlook in Section~\ref{sec:conclusions}.

\section{Measurements on Random Hyperplanes and Their Theory}
\label{sec:th}
To understand the nature of the objective landscape of neural networks, we constrain our optimization to a randomly chosen $d$-dimensional hyperplane in the full $D$-dimensional configuration space, similarly to \cite{intrinsic}. Due to the low dimension of our subspace $d \ll D$, we can evaluate the second derivatives of the loss function directly using automatic differentiation. The second derivatives, described by the \textit{Hessian} matrix $H \in \mathbb{R}^{d \times d}$, characterize the local convexity (or rather the amount of local curvature, as the function is not strictly convex, although we will use the terms convexity and curvature interchangeably throughout this paper) of the loss function at a given point. As such, they are a useful probe of the valley and hill-like nature of the local neighborhood of a given configuration, which in turn influences optimization.

\subsection{Random Hyperplanes}
\label{sec:eta}
We study the behavior of the loss function on random, low-dimensional hyperplanes, and later generalize to random, low-dimensional hyperspheres. Let the dimension of the full space be $D$ and the dimension of the hyperplane $d$. To specify a plane, we need $d$ orthogonal vectors $\{ \vec{v} \in \mathbb{R}^D \}$. The position within the plane is specified by $d$ coordinates, encapsulated in a position vector $\vec{\theta} \in \mathbb{R}^d$, as illustrated in Figure~\ref{fig:sphere}.
Given the origin of the hyperplane $\vec{P} \in \mathbb{R}^D$, the full-space coordinates $\vec{x}$ of a point specified by $\vec{\theta}$ are $\vec{x} = \vec{P} + \sum_{i=1}^d \theta_i \vec{v}_i$. This can be written simply as $\vec{x} \left ( \vec{\theta} \right ) = \vec{P} + M \vec{\theta}$, where $M \in \mathbb{R}^{D \times d}$ is a transformation matrix whose columns are the orthogonal vectors $\{ \vec{v} \}$. In our experiments, the initial $\vec{P}$ and $M$ are randomly chosen, and \textit{frozen} -- they are not trainable. To account for the different initialization radii of our weights, we map $M \to \eta M$, where $\eta_{ii} \propto 1/r_i$ is a diagonal matrix serving as a metric. The optimization affects solely the within-hyperplane coordinates $\vec{\theta}$, which are initialized at zero. The freedom to choose $\vec{P}$ according to modern initialization schemes allows us to explore realistic conditions similar to full-space optimization. 

As demonstrated in Section~\ref{sec:shell}, common initialization schemes choose points at an approximately fixed radius $r = |\vec{P}|$. In the low-dimensional hyperplane limit $d \ll D$, it is exceedingly unlikely that the hyperplane has a significant overlap with the radial direction $\hat{r}$, and we can therefore visualize the hyperplane as a \textit{tangent} plane at radius $|P|$.
This is illustrated in Figure~\ref{fig:sphere}.

For implementation reasons, we decided to generate \textit{sparse}, nearly-orthogonal projection matrices $M$ by choosing $d$ vectors, each having a random number $n$ of randomly placed non-zero entries, each being equally likely $\pm 1/\sqrt{n}$. Due to the low-dimensional regime $d \ll D$, such matrix is sufficiently near-orthogonal for our purposes, as validated numerically.
\begin{figure}[ht]
	\centering
	\begin{subfigure}{.23\textwidth}
		\centering
		\includegraphics[width=1.0\linewidth]{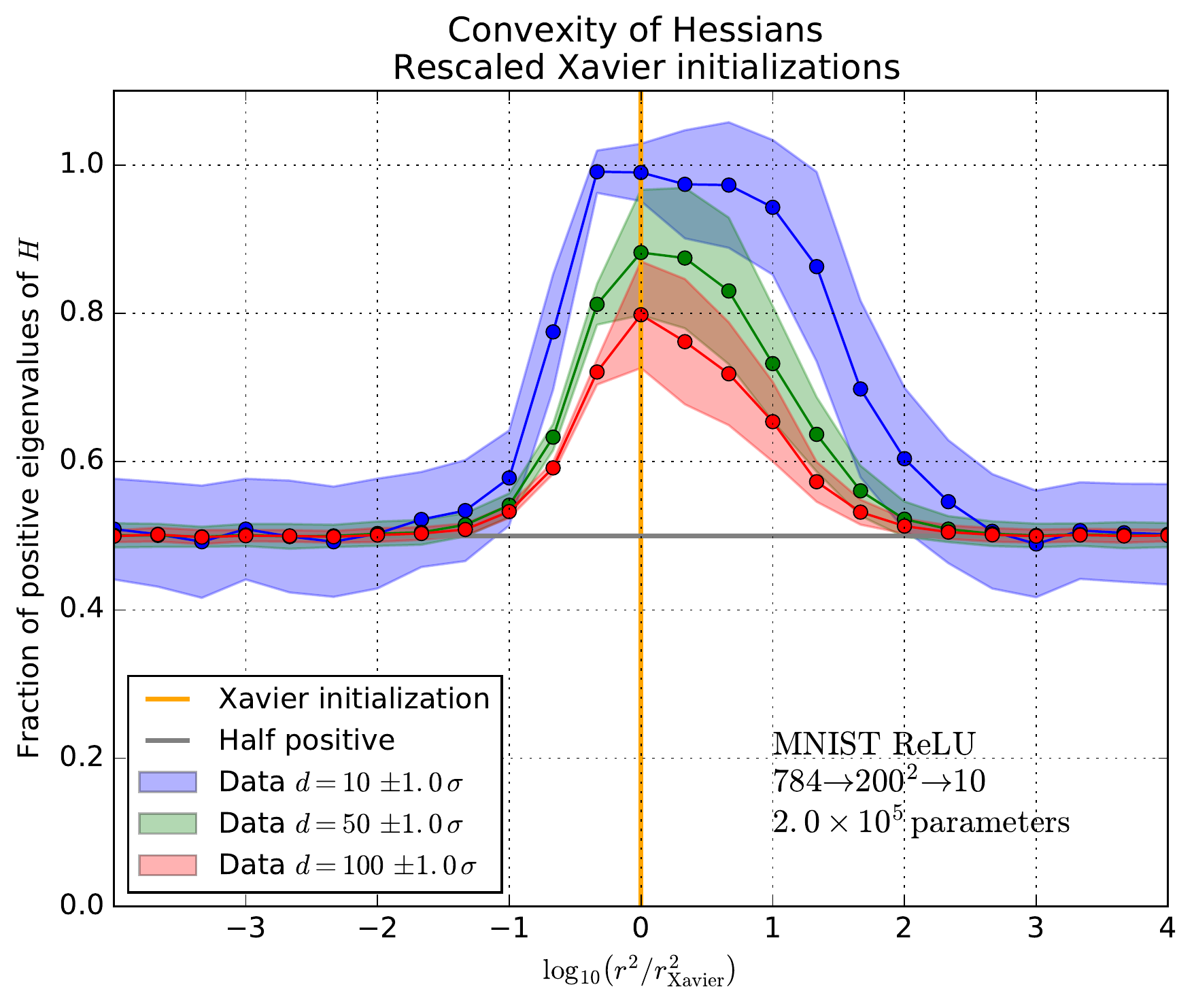}
		\caption{Fraction of positive eigenvalues.}
		\label{fig:fraction}
	\end{subfigure}\hfill
	\begin{subfigure}{.23\textwidth}
		\centering
		\includegraphics[width=1.0\linewidth]{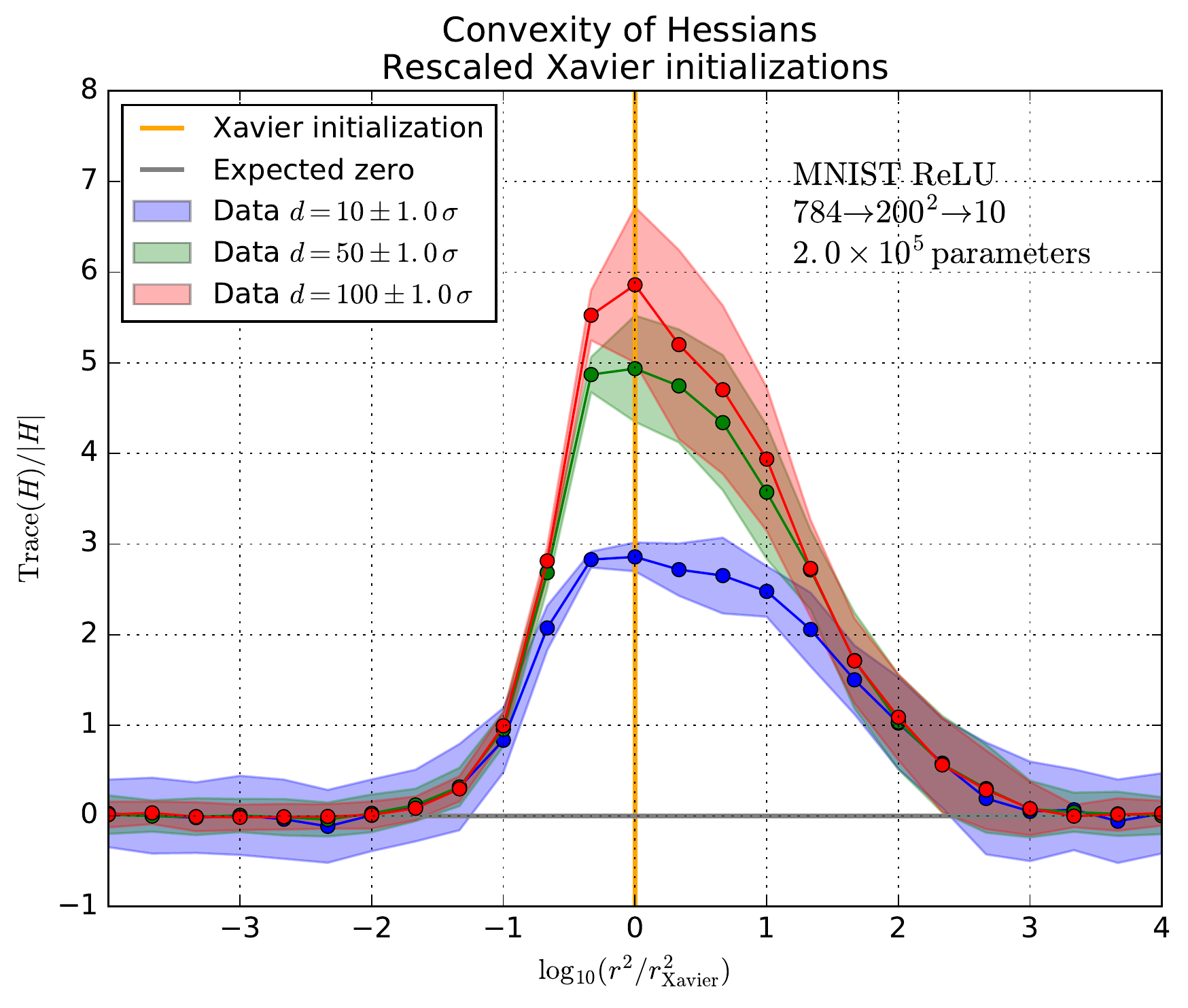}
		\caption{A measure of local convexity/positive curvature.}
		\label{fig:trHoverH}
	\end{subfigure}
	\caption{Two measures of local convexity/prevalence of positive curvature evaluated for random points using random, low-dimensional hyperplanes intersecting them. Figures~\ref{fig:fraction} and \ref{fig:trHoverH} show the existence of an unusual behavior of local convexity at a well defined range of configuration space radii -- the Goldilocks zone. The one sigma experimental uncertainties are shown as shading. The relatively small uncertainties in the Goldilocks zone point towards high angular isotropy of the objective landscape. The fraction of positive Hessian eigenvalues diminishes as the dimension $d$ of a random hyperplane increases, whereas $\mathrm{Tr}(H)/||H||$ remains a good predictor, as discussed theoretically in Section~\ref{sec:th}. The Xavier initialization \cite{Xavier} initializes networks.}
	\label{fig:sweep_single_archicture}
\end{figure}
\subsection{Gaussian Initializations on a Thin Spherical Shell}
\label{sec:shell}
Common initialization procedures \cite{Xavier,He} populate the network weight matrices $\{ W \}$ with elements drawn independently from a Gaussian distribution with mean $\mu = 0$ and a standard deviation $\sigma(W)$ dependent on the \textit{dimensionality} of the particular matrix. For a matrix $W$ with elements $\left \{ W_{ij} \right \}$, the probability of each element having value $w$ is $P(W_{ij} = w) \propto \exp\left( - \frac{w^2}{2 \sigma^2} \right)$.
The joint probability distribution is therefore
\begin{equation}
P(W) \propto \prod_{ij} \exp\left( - \frac{W^2_{ij}}{2 \sigma^2} \right) =  \exp\left( - \frac{r^2}{2 \sigma^2} \right ) \, ,
\end{equation}
where we define $r^2 \equiv \sum_{ij} W^2_{ij}$, i.e. the Euclidean norm where we treat each element of the matrix as a coordinate. The probability density at a radius $r$ therefore corresponds to
\begin{equation}
P(r) \propto r^{N-1} \exp\left( - \frac{r^2}{2 \sigma^2} \right ) \, ,
\end{equation}
where $N$ is the number of elements of the matrix $W$, i.e. the number of coordinates. The probability density for high $N$ peaks sharply at radius $r_* = \sqrt{N-1} \, \sigma$. That means that the bulk of the random initializations of $W$ will lie around this radius, and we can therefore visualize such random initialization as assigning a point to a thin spherical shell in the configuration space, as illustrated in Figure~\ref{fig:sphere}.

Each matrix $W$ is initialized according to its own dimensions and therefore ends up on a shell of a different radius in its respective coordinates. We compensate for this by introducing the metric $\eta$ discussed in Section~\ref{sec:eta}. Biases are initialized at zeros. There is a factor $\mathcal{O}(\sqrt D)$ fewer biases than weights in a typical fully-connected network, therefore we ignore biases in our theoretical treatment.
\subsection{Hessian}
We would like to explain the observed relationship between several sets of quantities involving second derivatives of the loss function on random, low-dimensional hyperplanes and hyperspheres. The loss at a point $\vec{x} = \vec{x_0} + \vec{\varepsilon}$ can be approximated as
\begin{equation}
L(\vec{x}_0 + \vec{\varepsilon}) = L(\vec{x_0}) + \vec g \cdot \vec{\varepsilon} + \frac{1}{2} \vec{\varepsilon}^T H \vec{\varepsilon} + \mathcal{O}(\varepsilon^3) \, .
\end{equation}
The first-order term involves the gradient, defined as $g_i = \partial L / \partial x_i$. The second-order term uses the second derivatives encapsulated in the Hessian matrix defined as $H_{ij} = \partial^2 L/\partial x_i\partial x_j$. The Hessian characterizes the \textit{local} curvature of the loss function. For a direction $\vec{v}$, $\vec{v}^T H \vec{v} > 0$ implies that the loss is \textit{convex} along that direction, and conversely $\vec{v}^T H \vec{v} < 0$ implies that it is concave.
We can diagonalize the Hessian matrix to its eigenbasis, in which its only non-zero components $\{h_i\}$ lie on its diagonal. We refer to its eigenvectors as the principal directions, and its eigenvalues $\{h_i\}$ as the principal curvatures.

After restricting the loss function to a $d$-dimensional hyperplane, we can compute a new Hessian, $H_d$, which comes with its own eigenbasis and eigenvalues. Each hyperplane, therefore, has its own set of principal directions and principal curvatures.

In this paper, we study the behavior of two quantities: $\mathrm{Tr}(H)$ and $||H||$. $\mathrm{Tr}(H)$ is the trace of the Hessian matrix and can be expressed as $\mathrm{Tr}(H) = \sum_i h_i$. As such, it can be thought of as the \textit{total} curvature at a point. $||H||^2 = \sum_i h_i^2$, and it is proportional to the \textit{variance} of curvature in different directions around a point (assuming large $D$ and several other assumptions, as discussed later).

\subsection{Curvature Statistics and Connections to Experiment}
As discussed in detail in Section~\ref{sec:results}, we observe that the following three things occur in the Goldilocks zone: 1) The curvature along the vast majority of randomly chosen directions is positive.
2) A slim majority of the principal curvatures (Hessian eigenvalues) $\{h_i\}$ is positive. If we restrict to a $d$-dimensional hyperplane, this majority becomes more significant for smaller $d$, as demonstrated in Figure \ref{fig:fraction}. 3) The quantity $\mathrm{Tr}(H)/||H||$ is significantly greater than 1 (see Figures~\ref{fig:bumps}, and \ref{fig:trHoverH}), and is close to $\mathrm{Tr}(H_d)/||H_d||$ for large enough $d$ (see Figure~\ref{fig:trHoverH}).

We can make sense of these observations using two tools: the fact that high-dimensional multivariate Gaussian distributions have correlation functions that approximate those of the uniform distribution on a hypersphere~\cite{MR2335894} and the fact (via Wick's Theorem) that for multivariate Gaussian random variables $(X_1,X_2,X_3,X_4)$,
\begin{eqnarray*}
E[X_1X_2X_3X_4]=E[X_1X_2]E[X_3X_4]+E[X_1X_3]E[X_2X_4]\\
+E[X_1X_4]E[X_2X_3]
\end{eqnarray*}

Choose a uniformly-random direction in the full $D$-dimensional space, given by a vector $v$. For $D\gg1$, any pair of distinct components $v_i,v_j$ are approximately bivariate Gaussian with $E[v_i]=0,E[v_i^2]=1/D,E[v_iv_j]=0$. Then, working in the eigenbasis of $H$, the expected curvature in the $v$ direction is $E[v^THv]=E\left[\sum_ih_i v_i^2\right]=\sum_ih_i/D=\mathrm{Tr}(H)/D$, which is the same as the average principal curvature. 

The variance of the curvature in the $v$-direction is
$$E[(v^THv)^2]-E[v^THv]^2=E\left[\sum_{ij}{h}_iv_i^2h_jv_j^2\right]-E[v^THv]^2$$
and we can apply Wick's Theorem to $E[v_iv_iv_jv_j]$ to find that this is $2\sum_ih_i^2/D^2=2||H||^2/D^2$. On the other hand, the variance of the principal curvatures is $(1/D)\sum_ih_i^2-((1/D)\sum_ih_i)^2=||H||^2/D-\mathrm{Tr}(H)^2/D^2$. Empirically, we find that $\mathrm{Tr}(H)/||H||\ll D$ in all cases we consider, so the dominant term is $||H||^2/D$, a factor of $D/2$ larger than we found for the $v$-direction. 

Therefore, when $D$ is large, the principal directions have the same average curvature that randomly-chosen directions do, but with much more variation. This explains why a slight excess of positive eigenvalues $h_i$ can correspond to an overwhelming majority of positive curvatures in random directions. 

In light of the calculations above, the condition $\mathrm{Tr}(H)/||H||\gg 1$ implies that the average curvature in a random direction is much greater than the standard deviation, i.e. that almost no directions have negative curvature. This is the hallmark of the Goldilocks zone, as demonstrated in Figure~\ref{fig:bumps}. The corresponding condition for principal directions to be overwhelmingly positive-curvature (i.e. for nearly all $h_i$ to be positive) is $\mathrm{Tr}(H)/||H|| \gg \sqrt{D}$, which is a much stronger requirement. Empirically, we find that in the Goldilocks zone, $\mathrm{Tr}(H)/||H||$ is greater than 1 but less than $\sqrt{D}$.

Restricting optimization to low-dimensional hyperplanes does not affect the ratio $\mathrm{Tr}(H)/||H||$. Another application of Wick's Theorem gives that $\mathrm{Tr}(H_d) \propto d$ for $d\gg 1$. Similarly, it can be shown that $|H_d| \propto d$ so long as $d\gg \left(\mathrm{Tr}(H)/||H||\right)^2$. This implies that for large enough $d$, $\mathrm{Tr}(H_d)/||H_d||\approx \mathrm{Tr}(H)/||H||$. We find empirically that at $r_{\mathrm{Xavier}}$ this ratio begins to stabilize for $d$ greater than about $\left(\mathrm{Tr}(H)/||H||\right)^2\approx 40$, as shown in Figure~\ref{fig:trHoverH}.

Because $||H_d||\propto d$, the principal curvatures on $d$-hyperplanes have standard deviation $||H_d||/\sqrt{d} \propto \sqrt{d}$, while the average principal curvature stays constant. This explains why smaller $d$ hyperplanes have a larger excess of positive eigenvalues in the Goldilocks zone than larger $d$ hyperplanes (Figure \ref{fig:fraction}): their principal curvatures have the same (positive) average value, but vary less, on hyperplanes of smaller $d$.

As a final observation, we note that the beginning and end of the Goldilocks zone occur for different reasons. The increase in $\mathrm{Tr}(H)/||H||$ at its inner edge comes from an increase in $\mathrm{Tr}(H)$, while the decrease at its outer edge comes from an increase in $||H||$, as shown in Figure \ref{fig:trHrabsHr}. To (partially) explain this, we give a more detailed analysis of these two quantities below. It turns out that the increase in $\mathrm{Tr}(H)$ is a feature of the network architecture, while the decrease in $||H||$ is not as well understood and may be a feature of the problem domain.

\subsection{Radial Dependence of Loss}
We observe that for fully-connected neural networks with the $\mathrm{ReLU}$ non-linearity and a final $\mathrm{softmax}$, the loss at an average configuration at distance $r$ from the origin is constant for $r \lessapprox r_\mathrm{Goldilocks}$, and grows as a power-law with the exponent being the number of layers for $r \gtrapprox r_\mathrm{Goldilocks}$, as shown in Figure~\ref{fig:loss_scaling}. To connect these observations to theory, we note that, for $n_L$ layers (i.e. $n_L-1$ hidden layers), each numerical output gets matrix-multiplied by $n_L$ layers of weights. If all weights are rescaled by some number $r$, 1) the radius of configuration space position grows by the factor $r$, and 2) the raw output from the network will generically grow as $r^{n_L}$ due to the linear regime of the $\mathrm{ReLU}$ activation function.

Suppose that the raw outputs from the network (before $\mathrm{softmax}$) are order from the largest to the smallest as $(y_1,y_2,\dots)$. The difference $\Delta y$ between any two outputs $y_i$ and $y_j$ ($i<j$) will grow as $r^{n_L}$. After the application of $\mathrm{softmax}$ $p_1 =1 - \exp(-r^{n_L} (y_1-y_2))$ and $p_2 = \exp(-r^{n_L} (y_1-y_2))$.
If the largest output $p_1$ corresponds to the correct class, its cross entropy contribution will be $-\log(p_1) \approx \exp(-r^{n_L} (y_1-y_2)) \to 0$. However, if instead $p_2$ corresponds to the correct answer, as expected at 10 \% of the cases for a random network on MNIST or CIFAR-10, the cross-entropy loss contribution is $-\log(p_2) \approx r^{n_L} (y_1-y_2) \propto r^{n_L}$. This demonstrates that at large configuration space radii $r$, a small, random fluctuation in the raw outputs of the network will have an exponential influence on the resulting loss. On the other hand, if $r$ is small, the $\mathrm{softmax}$ will bring the outputs close to one another and the loss will be roughly constant (equal to $\log (10)$ for 10-way classification).

Therefore, above some critical $r$, the loss will grow as a power law with exponent $n_L$. Below this critical value, it will be nearly flat. We confirm this empirically, as shown in Figure~\ref{fig:loss_scaling}. This scaling of the loss does not apply to $\tanh$ activations, 
which have a much more bounded linear regime and do not produce arbitrarily large outputs, which might explain the weaker presence of the Goldilocks zone for them, as seen in Figure~\ref{fig:bumps}.

\subsection{Radial Features and the Laplacian}
As discussed above, $\mathrm{Tr}(H)$ measures the excess of positive curvature at a point. It is also known as the Laplacian, $\nabla^2L$. Large positive values occur at valley-like points: not necessarily local minima, but places where the positively curved directions overwhelm the negatively curved ones. Similarly, very negative values occur at hill-like points, including local maxima. This intuition explains our observation that, in the Goldilocks zone, $\mathrm{Tr}(H)/||H||$ is strongly negatively correlated with the loss, as demonstrated in Figure~\ref{fig:correlation}. However, it is uncorrelated with accuracy, suggesting that $\mathrm{Tr}(H)/||H||$ indicated a point's suitability for optimization rather than its suitability as a final point. 

This leads to a puzzle, however. If we consider all of the points on a sphere of some radius, then a positive-on-average $\mathrm{Tr}(H)$ corresponds to an excess of valleys over hills.
On the other hand, one might expect hills and valleys to cancel out for a function on a compact manifold like a sphere.
Yet we empirically observe such an excess, as shown in Figure~\ref{fig:bumps}.

The explanation of this puzzles relies on our observation that the average loss begins to increase sharply at the inner edge of the Goldilocks zone (see Figure~\ref{fig:loss_scaling}). A line tangent to the sphere (perpendicular to the $\hat{r}$ direction) will head out to larger $r$ in both directions, therefore if the loss is increasing with radius, the point of tangency will be approximately at a local minimum. This means that every point -- whether it is a local maximum, a local minimum, or somewhere in between on the sphere of $r=\mathrm{const.}$ -- has a Hessian ($\propto$ Laplacian) that looks a bit more valley-like than it otherwise would, provided that we measure it within a low-dimensional hyperplane containing said point.

To be more precise, it follows from Stokes' Theorem that the average value of $\mathrm{Tr}(H)$ over a sphere (i.e. the surface of $r=\mathrm{const}.$) is given by
\begin{equation}
\fontsize{6.5pt}{0.0}
\langle\mathrm{Tr}(H)\rangle_{r = \mathrm{const.}} = \frac{\partial^2}{\partial r^2}\langle L\rangle_{r = \mathrm{const.}} + \frac{D-1}r\frac{\partial}{\partial r}\langle L\rangle_{r = \mathrm{const.}} \, ,
\end{equation}
where $\langle\cdot\rangle_{r = \mathrm{const.}}$ indicates averaging over a spherical surface of a constant radius. Therefore, this radial variation is the only thing that can \textit{consistently} affect $\mathrm{Tr}(H)$ in a specific way -- the other contributions, from the actual variation in the loss over the sphere, necessarily average out to zero, regardless of the specific form of $L$. The true hills and valleys cancel out, and the fake valleys from the radial growth remain.

We can verify this picture by showing that the $r$-dependence of $\mathrm{Tr}(H)$ is consistent with that of $\langle L\rangle_{r = \mathrm{const.}}(r)$. For a network with two hidden layers, $L(r)\propto r^{n_L} = r^3$ for large $r$, as shown above. Therefore, $\frac{D-1}r\frac{\partial L}{\partial r}$ grows linearly in $r$. Consistent with this, we find that $\mathrm{Tr}(H)/r$ is constant for large $r$, as shown in Figure~\ref{fig:trHrabsHr}.

We observe that in the Goldilocks zone $\mathrm{Tr}(H)$ is a fairly consistent function of radius, with small fluctuations around its average, which indicates that this fake-valley effect dominates there: $\mathrm{Tr}(H)$, and therefore the average curvature, is controlled mostly by the radial feature we observe, rather than anisotropy of the loss function on the sphere (i.e. in the angular directions).

The increase in loss, as explained above in terms of the properties of weight matrices and cross-entropy, shows that the increase in $\mathrm{Tr}(H)$ (and therefore the inner edge of the Goldilocks zone) is caused by the same effect that is usually invoked to explain why common initialization techniques work well -- prevention of exponential explosion of signals. In particular, it explains why the inner edge of the zone is close to the prescribed $r_*$ of these initialization techniques.

On the other hand, $||H||$ is not nearly as strongly affected by the radial growth. This is related to the variance in curvature, not the average curvature; every tangent direction gets the same fake-valley contribution, so $||H||$ is mostly unaffected. 
(To be more precise, this is true as long as the principal curvatures have a mean smaller than their standard deviation, $\mathrm{Tr}(H)/||H||\ll \sqrt{D}$, which is true for all of our experiments.)
While $\mathrm{Tr}(H)$ is \textit{global} and \textit{radial}, $||H||$ is \textit{local} and \textit{angular}: it is mostly determined by how the loss varies in the $D-1$ angular directions along a sphere of $r = \mathrm{const}$. This means that the sudden increase in $||H||$ (and the outer edge of the Goldilocks zone) seen in Figure \ref{fig:trHrabsHr} is due to an increase in the \textit{texture}-like angular features of the loss, the sharpness of its hills and valleys, past some value of $r$. This will require much more research to understand in detail, and is beyond the scope of this paper.

\section{Results and Discussion}
\label{sec:results}
\begin{figure}[ht]
	\centering
	\begin{subfigure}{.23\textwidth}
		\centering
		\includegraphics[width=1.0\linewidth]{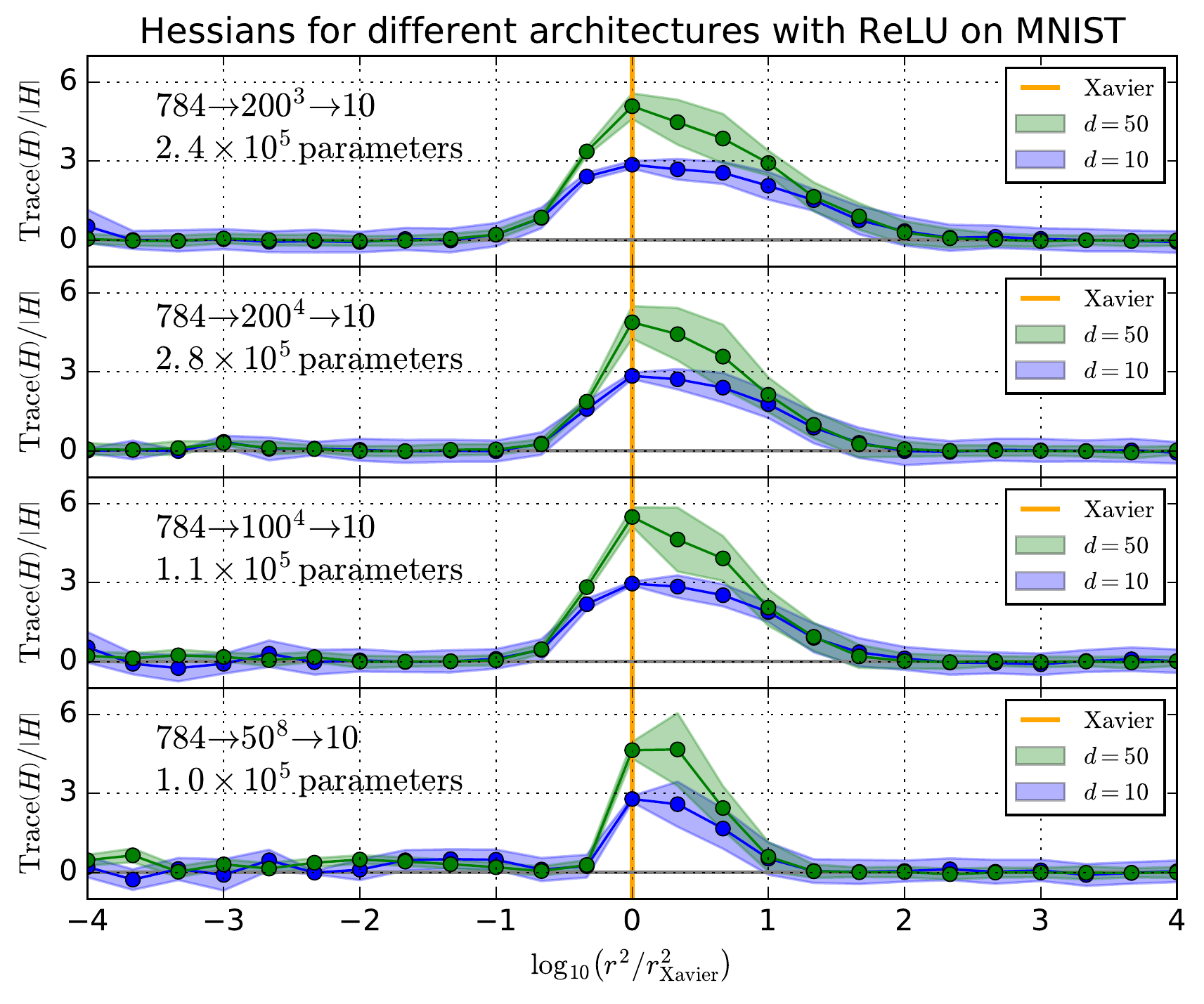}
		\caption{Fully-connected NNs with $\mathrm{ReLU}$ on MNIST}
		\label{fig:MNIST}
	\end{subfigure}\hfill
	\begin{subfigure}{.23\textwidth}
		\centering
		\includegraphics[width=1.0\linewidth]{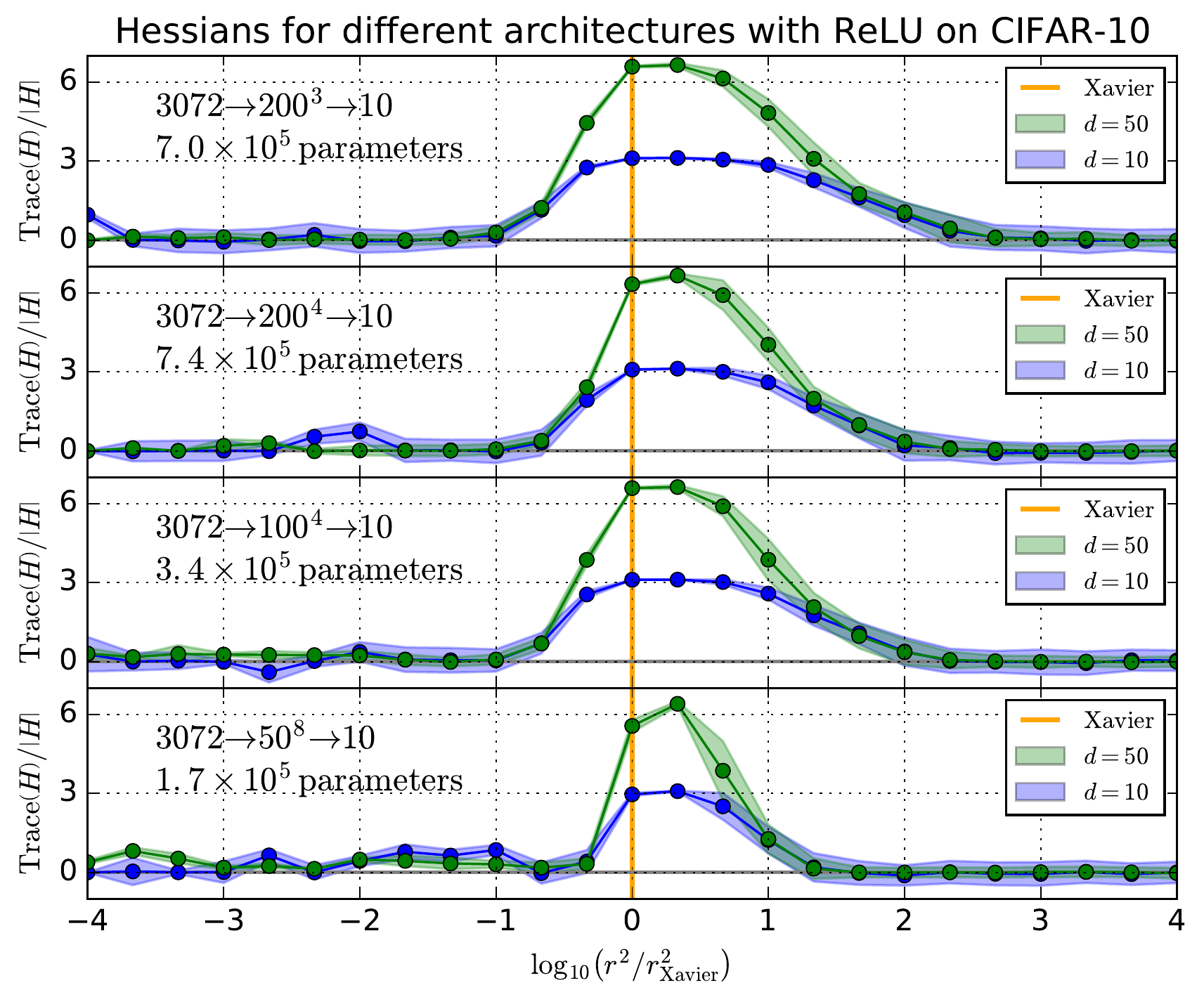}
		\caption{Fully-connected NNs with $\mathrm{ReLU}$ on CIFAR-10}
		\label{fig:CIFAR}
	\end{subfigure}
	
	\bigskip
	\begin{subfigure}{.23\textwidth}
		\centering
		\includegraphics[width=1.0\linewidth]{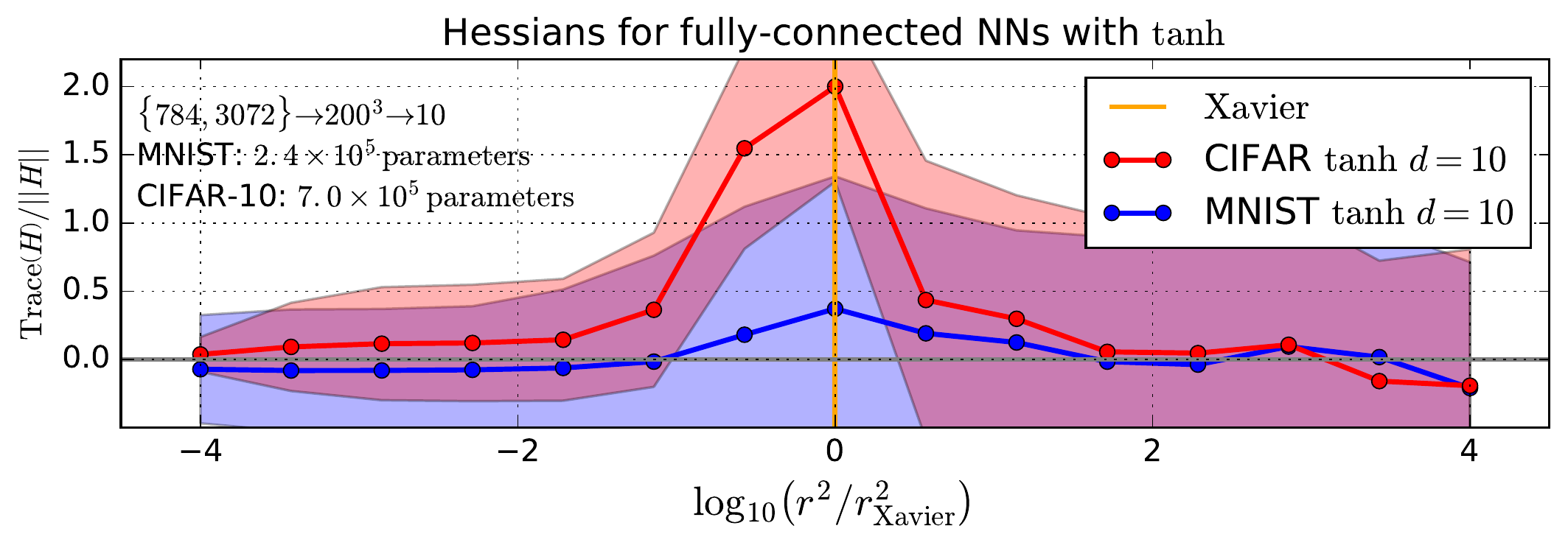}
		\caption{Fully-connected NNs with $\mathrm{tanh}$ on MNIST and CIFAR-10}
		\label{fig:MNIST_tanh}
	\end{subfigure}\hfill
	\begin{subfigure}{.23\textwidth}
		\centering
		\includegraphics[width=1.0\linewidth]{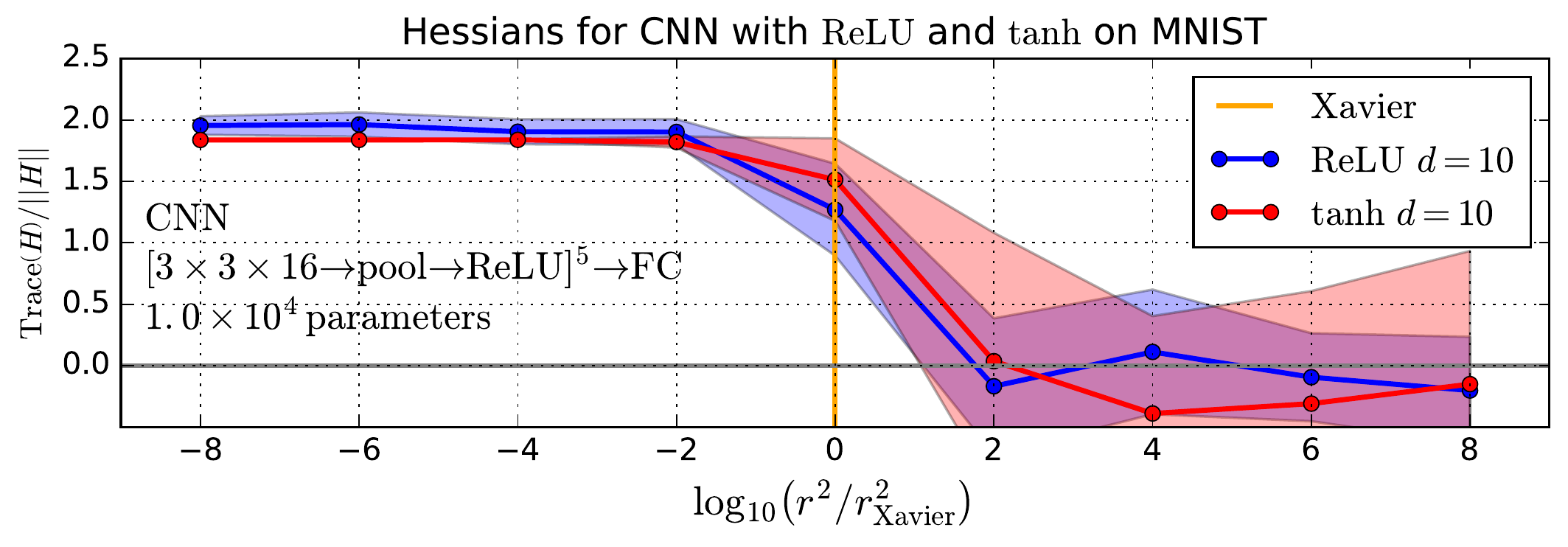}
		\caption{CNNs with $\mathrm{ReLU}$ and $\mathrm{tanh}$ on MNIST}
		\label{fig:MNIST_tanh}
	\end{subfigure}
	\caption{Characteristics of Hessians at randomly initialized points on random hyperplanes at different radii. The plots show $\mathrm{Tr}(H)/||H||$ which is a good tracer of the amount of local convexity/prevalence of positive curvature, as discussed in Section~\ref{sec:th}. The effect appears consistently for the MNIST and CIFAR-10 datasets, a range of fully-connected network widths and depths, as well as the $\mathrm{ReLU}$ and $\tanh$ non-linearities. The peak coincides with the radius on which the Xavier scheme initializes neural networks, suggesting a link between the local convexity and signal growth. For CNNs, a different profile of convexity is observed, although its unusual non-zero size remains.}
	\label{fig:bumps}
\end{figure}

\begin{figure}[ht]
	\centering
	\begin{subfigure}{.22\textwidth}
		\centering
		\includegraphics[width=1.0\linewidth]{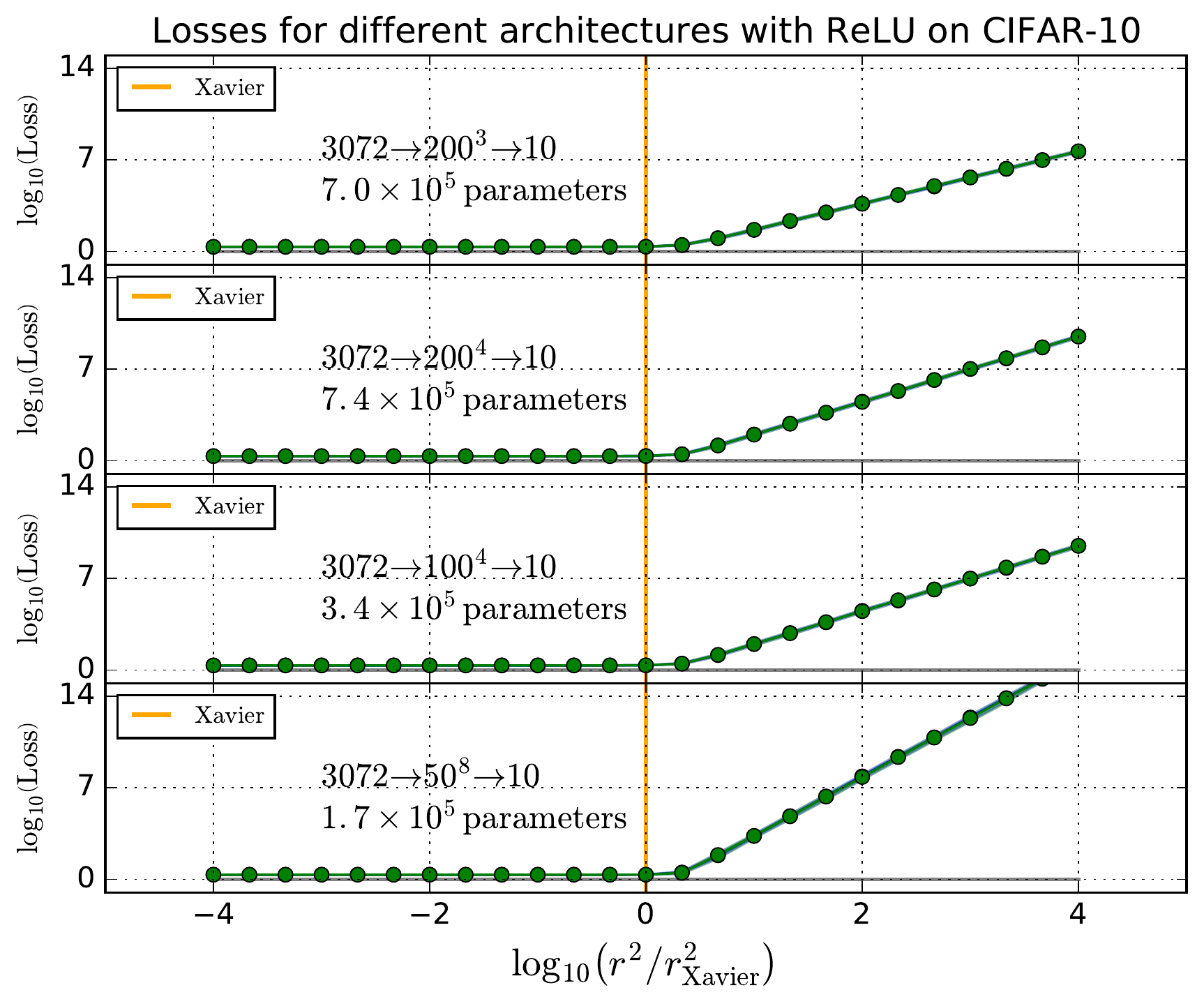}
		\caption{The plot shows the scaling of the average loss at randomly initialized points as a function of radius for 4 different architectures. The loss is approximately constant for $r \lessapprox r_\mathrm{Xavier}$ and power-law with the exponent predicted in Section~\ref{sec:th} for $r \gtrapprox r_\mathrm{Xavier}$.}
		\label{fig:loss_scaling}
	\end{subfigure}\hfill
	\begin{subfigure}{.22\textwidth}
		\centering
		\includegraphics[width=1.0\linewidth]{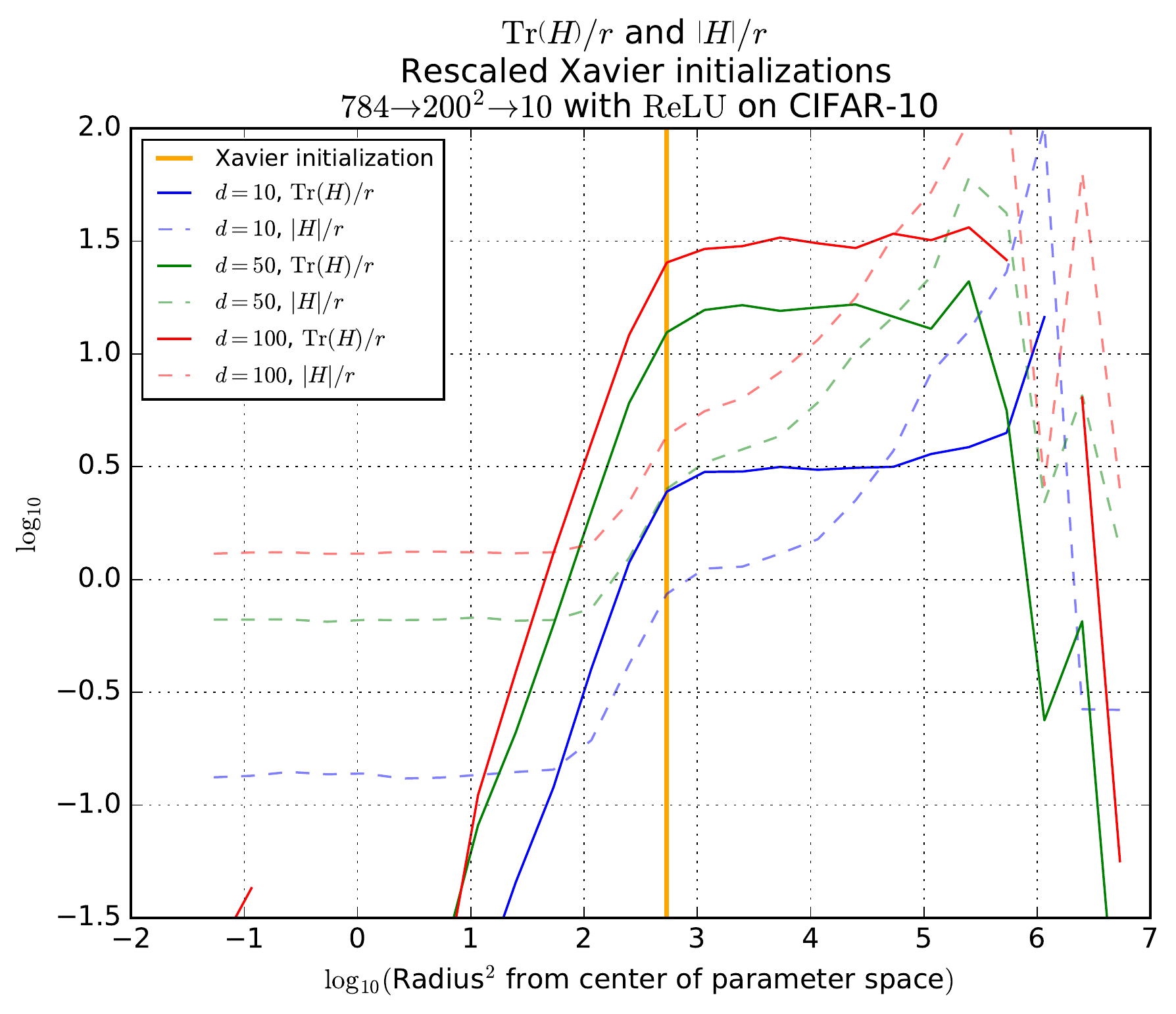}
		\caption{The plot shows the $\mathrm{Trace}(H)/r$ and $||H||/r$ for the cross-entropy loss Hessian as a function of radius for 3 different hyperplane dimensionalities. The intersections between the $\mathrm{Trace}(H)$ and $||H||$ curves correspond to the Goldilocks zone boundaries.}
		\label{fig:trHrabsHr}
	\end{subfigure}
	\caption{Properties of the cross-entropy loss function for a fully-connected network with  $\mathrm{ReLU}$.}
	\label{fig:props}
\end{figure}

\begin{figure}[ht]
	\centering
	\begin{subfigure}[t]{.23\textwidth}
		\centering
		\includegraphics[width=1.0\linewidth]{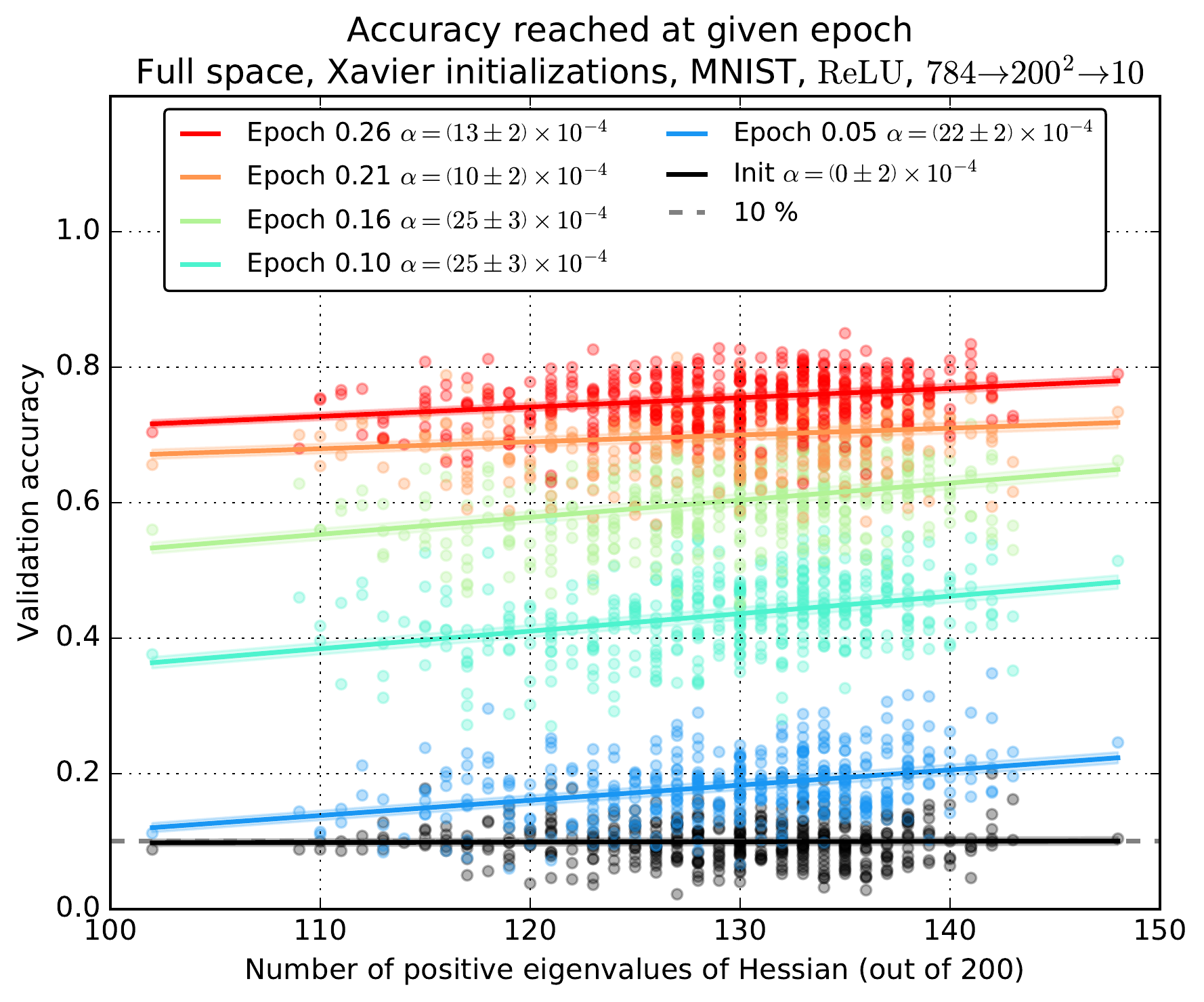}
		\caption{Accuracy vs. the number of positive Hessian eigenvalues.}
		\label{fig:acc_positive_evals}
	\end{subfigure}\hfill
	\begin{subfigure}[t]{.23\textwidth}
		\centering
		\includegraphics[width=1.0\linewidth]{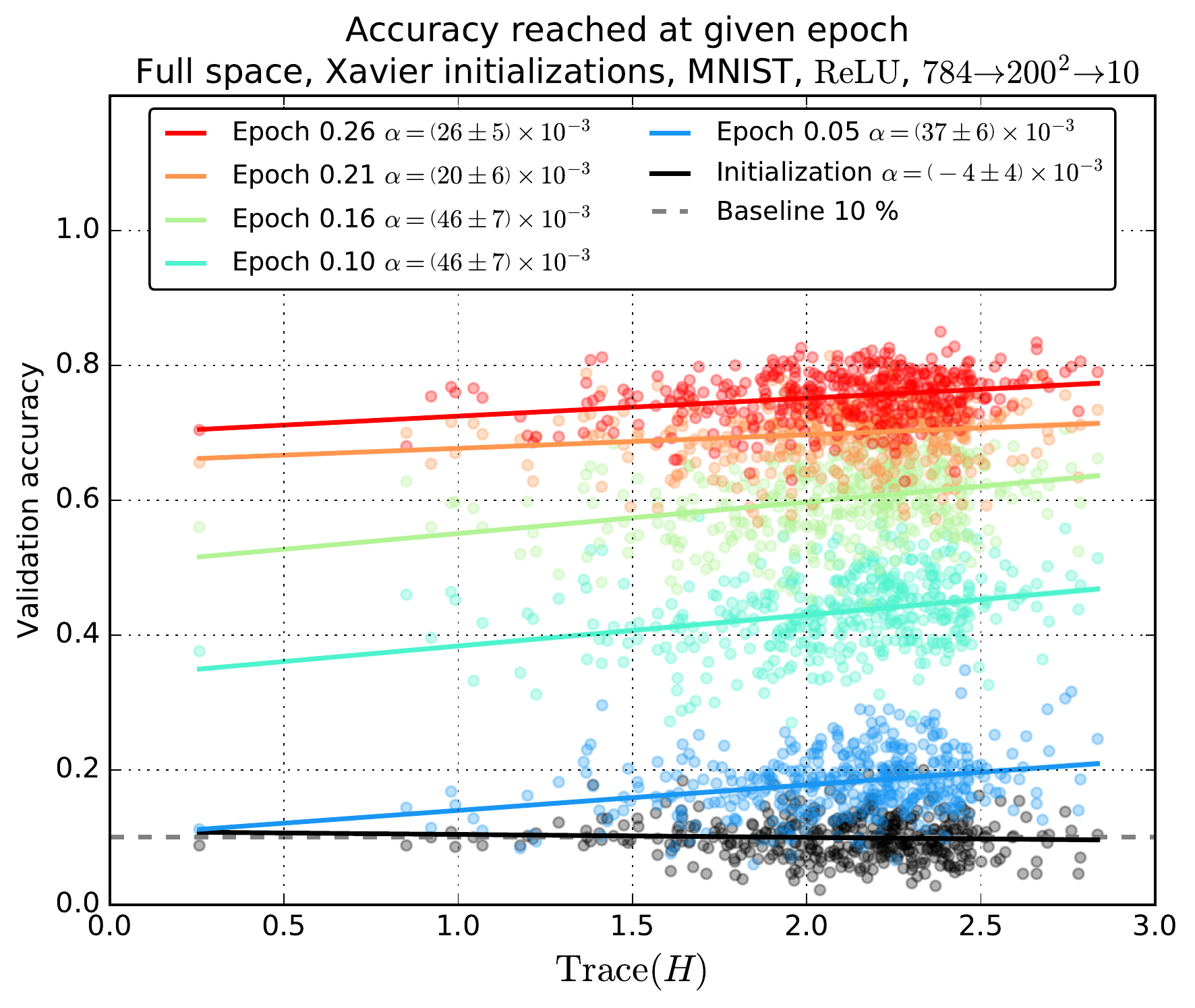}
		\caption{Accuracy vs. the initial trace of Hessian.}
		\label{fig:accs_trace}
	\end{subfigure}
	%
	\begin{subfigure}[t]{.23\textwidth}
		\centering
		\includegraphics[width=1.0\linewidth]{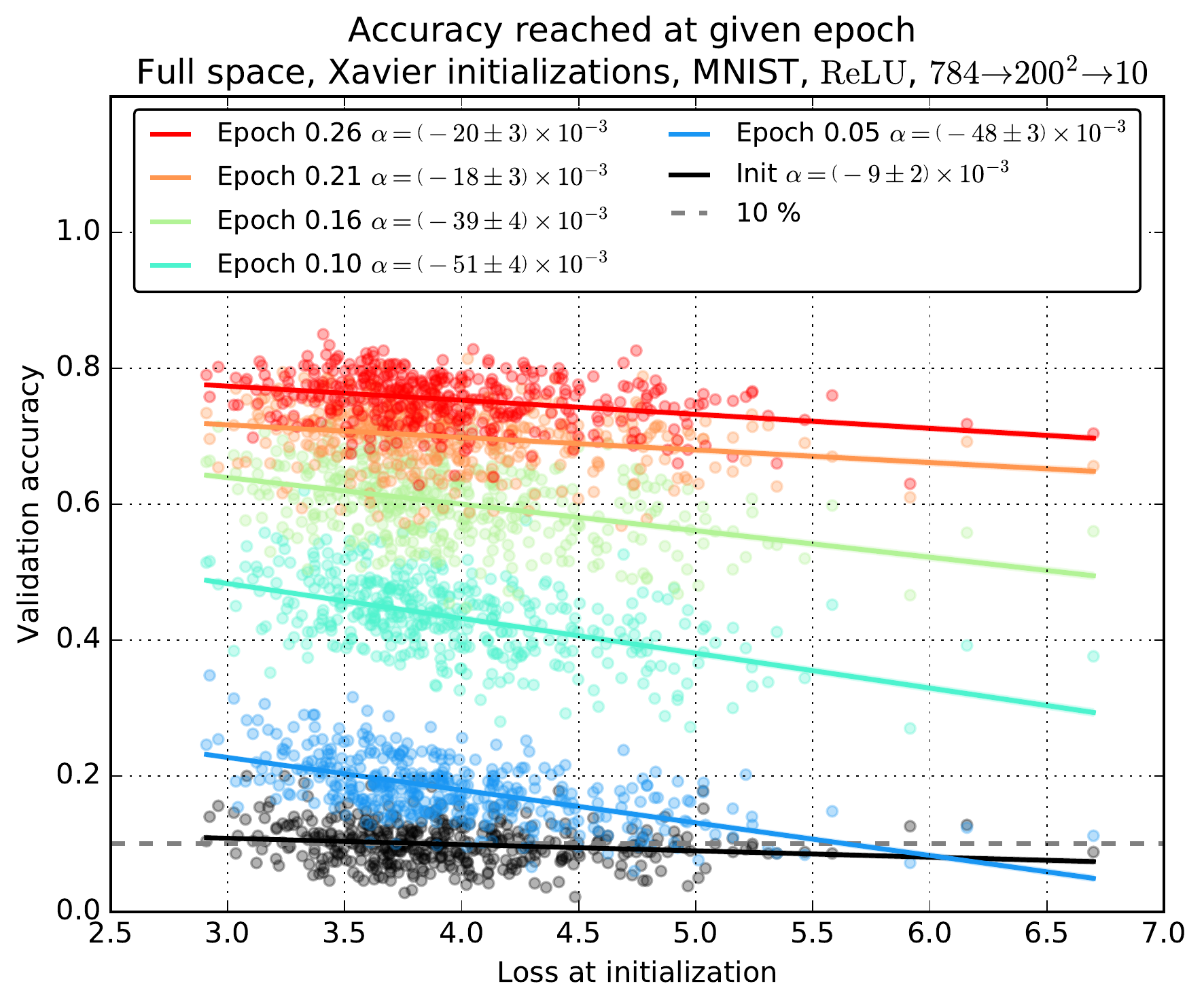}
		\caption{Accuracy vs. the initial validation loss.}
		\label{fig:accs_ini_loss}
	\end{subfigure}\hfill
	\begin{subfigure}[t]{.23\textwidth}
		\centering
		\includegraphics[width=1.0\linewidth]{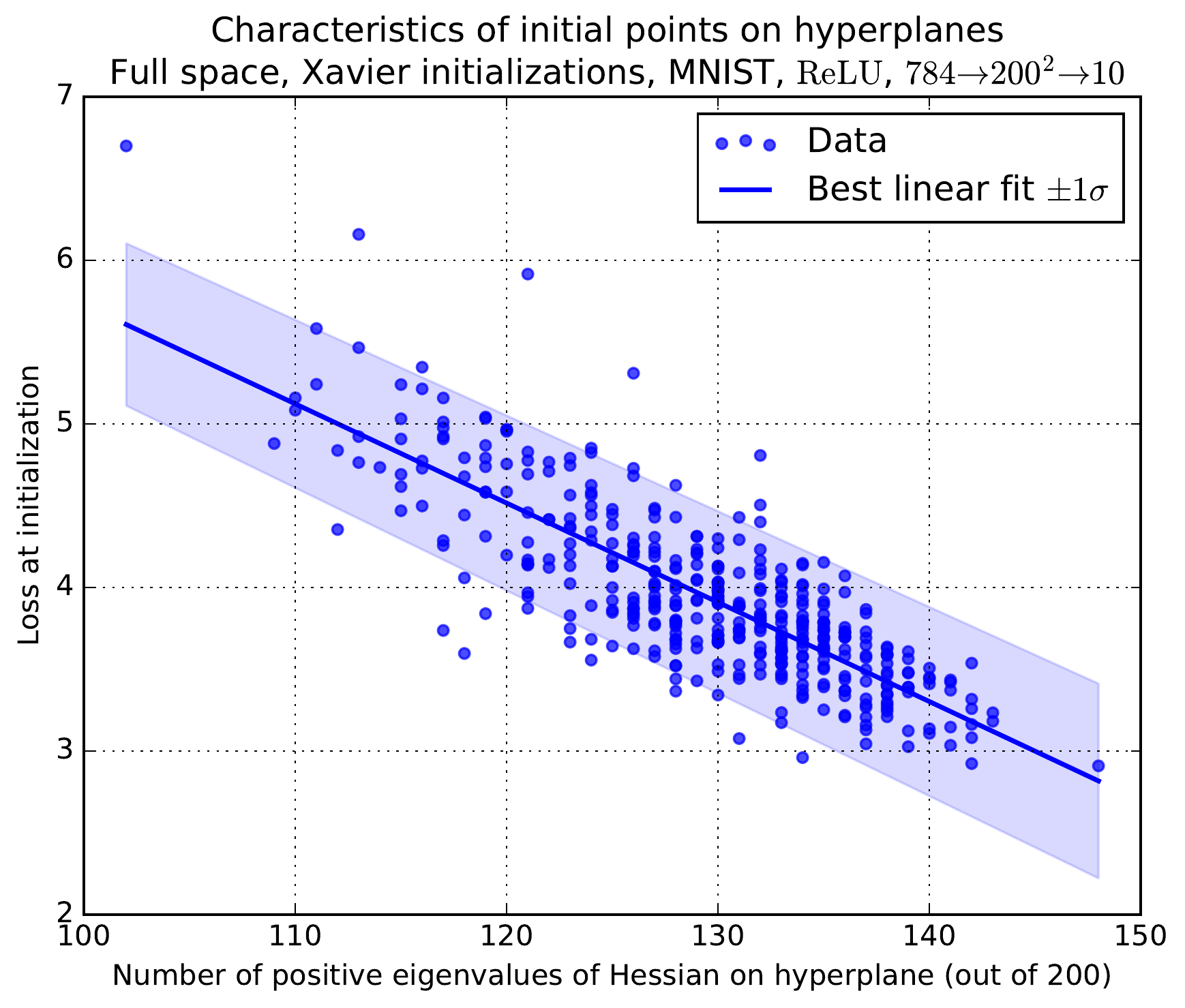}
		\caption{Initial validation loss vs. the number of positive eigenvalues.}
		\label{fig:correlation}
	\end{subfigure}
	\caption{Correlation between properties of Hessians at random initial points and the speed of optimization. Accuracies reached at a given epoch for random initial points with different Hessian properties are shown. The best fit line is plotted for each time and its slope is presented. The higher the trace of Hessian or the number of its positive eigenvalues ($\propto$ convexity), the faster the optimization. Due to the strong correlation between these three properties in the Goldilocks zone illustrated in panel d), sampling a number of initializations and choosing the lowest initialization loss, although unrelated to the initial accuracy, leads to faster convergence.}
	\label{fig:accs}
\end{figure}

We ran a large number of experiments to determine the nature of the objective landscape. We focused on fully-connected and convolutional networks. We used the MNIST \cite{lecun-mnisthandwrittendigit-2010} and CIFAR-10 \cite{CIFAR10} image classification datasets. We explored a range of widths and depths of fully-connected networks to determine the stability of our results, and considered the $\mathrm{ReLU}$ and $\tanh$ non-linearities. We used the cross-entropy loss and the Adam optimizer with the learning rate of $10^{-3}$.
We studied the properties of Hessians characterizing the local convexity/prevalence of positive curvature of the loss landscape on randomly chosen hyperplanes. We observed the following:
\begin{enumerate}
	\item An unusually high fraction ($>1/2$) of positive eigenvalues of the Hessian at randomly initialized points on randomly oriented, low-dimensional hyperplanes intersecting them. The fraction increased as we optimized within the respective hyperplanes, and decreased with an increasing dimension of the hyperplane, as predicted in Section~\ref{sec:th}. The effect appeared at a well defined range of coordinate space radii we refer to as the \textit{Goldilocks zone}, as shown in Figure~\ref{fig:fraction} and illustrated in Figure~\ref{fig:sphere}.
	\item An unusual, statistically significant excess of $\mathrm{Tr}(H) / ||H||$ at the same region, as shown in Figures~\ref{fig:MNIST} and \ref{fig:CIFAR}. Unlike the excess of positive Hessian eigenvalues, the effect did not decrease with an increasing dimension of the hyperplane and was therefore a good tracer of local convexity/prevalence of positive curvature. Theoretical justification is provided in Section~\ref{sec:th}, and scaling with $d$ is shown in Figure~\ref{fig:trHoverH}.
	\item The observations 1. and 2. were made consistently over different widths and depths of fully-connected neural networks on MNIST and CIFAR-10 using the $\mathrm{ReLU}$ and $\tanh$ non-linearities (see Figure~\ref{fig:bumps}).
	\item The loss function at random points at radius $r$ from the origin is \textit{constant} for $r \lessapprox r_\mathrm{Goldilocks}$ and grows as a power-law with the exponent predicted in Section~\ref{sec:th} for $r \gtrapprox r_\mathrm{Goldilocks}$, as shown in Figure~\ref{fig:loss_scaling}. The existence of the Goldilocks zone depends on the behavior of $\mathrm{Tr}(H)$ and $||H||$ shown in Figure~\ref{fig:trHrabsHr}.
	\item The accuracy reached on random, low-dimensional hyper\textit{planes} was good for $r<r_{\mathrm{Xavier}}$ and dropped dramatically for $r>r_{\mathrm{Xavier}}$, as shown in Figure~\ref{fig:accs_contours}. The accuracy reached on random, low-dimensional hyper\textit{spheres} was good only for $r \approx r_{\mathrm{Xavier}}$ and dropped for both $r<r_{\mathrm{Xavier}}$ and $r>r_{\mathrm{Xavier}}$, as shown in Figure~\ref{fig:accs_contours_spheres}.
	\item Common initialization schemes (such as Xavier \cite{Xavier} and He \cite{He}), initialize neural networks well within the Goldilocks zone, precisely at the radius at which measures of local convexity/prevalence of positive curvature peak (see Figures~\ref{fig:sweep_single_archicture}~and~\ref{fig:bumps}).
	\item Hints that selecting initialization points for high measures of local convexity leads to statistically significantly faster convergence (see Figure~\ref{fig:accs}), and correlates well with low initial loss. 
	\item Initializing at $r<r_\mathrm{Goldilocks}$, full-space optimization draws points to $r \approx r_\mathrm{Goldilocks}$. This suggests that the zone contains a large amount of suitable final points as well as suitable initialization points.
\end{enumerate}
An illustration of the Goldilocks zone and its relationship to the common initialization radius is shown in Figure~\ref{fig:sphere}.

We sampled a large number of Xavier-initialized points, used random hyperplanes intersecting them to evaluate their Hessians, and optimized within the \textit{full $D$-dimensional space} for 10 epochs starting there. We observe hints that a) the higher the fraction of positive eigenvalues, and b) the higher the $\mathrm{Trace}(H)$, the faster our network reaches a given accuracy on MNIST. Our experiments are summarized in Figures~\ref{fig:acc_positive_evals}, ~\ref{fig:accs_trace} and ~\ref{fig:accs_ini_loss}. We observe that the initial validation loss for Xavier-initialized points negatively correlates with these measures of convexity, as shown in Figure~\ref{fig:correlation}. This suggests that selecting for a low initial loss, even though uncorrelated with the initial accuracy, leads to faster convergence.

Using our observations, we draw the following conclusions:
\begin{enumerate}
	\item There exists a thick, hollow, spherical shell of unusually high local convexity/prevalence of positive curvature we refer to as the \textit{Goldilocks zone} (see Figure~\ref{fig:sphere} for an illustration, and Figures~\ref{fig:sweep_single_archicture} and \ref{fig:bumps} for experimental data).
	\item Its existence comes about due to an interplay between the behavior of $\mathrm{Tr}(H)$ and $||H||$, which we discuss in Section~\ref{sec:th}, and verify empirically (see Figures~\ref{fig:loss_scaling} and \ref{fig:trHrabsHr}).
	\item When optimizing on a random, low-dimensional hypersurface of dimensionality $d$, the overlap between the Goldilocks zone and the hypersurface is the main predictor of final accuracy reached, as demonstrated in Figure~\ref{fig:contours}.
	\item As a consequence, we show that the concept of \textit{intrinsic dimension} of a task introduced in \cite{intrinsic} is necessarily radius-dependent, and therefore initialization-dependent. Our results extend the results in \cite{intrinsic} to $r \neq r_\mathrm{Xavier}$ and to hyperspherical surfaces.
	\item The small variance between final accuracy reached by optimization constrained to random, low-dimensional hyperplanes is related to the hyperplanes being a) normal to $\hat{r}$, b) well within the Goldilocks zone, and c) the Goldilocks zone being angularly very isotropic.
	\item Hints that using a good initialization scheme and selecting for high measures of local convexity such as the number of positive Hessian eigenvalues, $\mathrm{Trace}(H)/||H||$, or low initial validation loss (they are all correlated (see Figure~\ref{fig:correlation}) in the Goldilocks zone), leads to faster convergence (see Figure~\ref{fig:accs}).
	\item Common initialization schemes (such as Xavier \cite{Xavier} and He \cite{He}), initialize neural networks well within the Goldilocks zone, consistently matching the radius at which measures of convexity peak. 
\end{enumerate}
Since measures of local convexity/prevalence of positive curvature peak in the Goldilocks zone, the intersection with the zone predicts optimization success in challenging conditions (constrained to a $(d \ll D)$-hypersurface), common initialization techniques initialize there, and points from $r<r_\mathrm{Goldilocks}$ are drawn there, we hypothesize that the Goldilocks zone contains an exceptionally high density of suitable initialization points as well as final points.
\section{Conclusion}
\label{sec:conclusions}
We explore the loss landscape of fully-connected and convolutional neural networks using random, low-dimensional hyperplanes and hyperspheres. We observe an unusual behavior of the second derivatives of the loss function -- an excess of local convexity -- localized in a range of configuration space radii we call the \textit{Goldilocks zone}. We observe this effect strongly for a range of fully-connected architectures with $\mathrm{ReLU}$ and $\tanh$ non-linearities on MNIST and CIFAR-10, and a similar effect for convolutional neural networks. We show that when optimizing on low-dimensional surfaces, the main predictor of success is the overlap between the said surface and the Goldilocks zone. We demonstrate connections to common initialization schemes, and show hints that local convexity of an initialization is predictive of training speed. We extend the analysis in \cite{intrinsic} and show that the concept of \textit{intrinsic dimension} is initialization-dependent. Based on our experiments, we conjecture that the Goldilocks zone contains high density of suitable initialization points as well as final points. We offer theoretical justifications for many of our observations.

\subsubsection*{Acknowledgments}
We would like to thank Yihui Quek and Geoff Penington from Stanford University for useful discussions.

\bibliography{goldilocks_biblio}

\begin{thebibliography}{}

\bibitem[\protect\citeauthoryear{Choromanska \bgroup et al\mbox.\egroup
  }{2014}]{DBLP:journals/corr/ChoromanskaHMAL14}
Choromanska, A.; Henaff, M.; Mathieu, M.; Arous, G.~B.; and LeCun, Y.
\newblock 2014.
\newblock The loss surface of multilayer networks.
\newblock {\em CoRR} abs/1412.0233.

\bibitem[\protect\citeauthoryear{Glorot and Bengio}{2010}]{Xavier}
Glorot, X., and Bengio, Y.
\newblock 2010.
\newblock Understanding the difficulty of training deep feedforward neural
  networks.
\newblock In {\em JMLR W\&CP: Proceedings of the Thirteenth International
  Conference on Artificial Intelligence and Statistics (AISTATS 2010)},
  volume~9,  249--256.

\bibitem[\protect\citeauthoryear{Goodfellow and
  Vinyals}{2014}]{line_goodfellow}
Goodfellow, I.~J., and Vinyals, O.
\newblock 2014.
\newblock Qualitatively characterizing neural network optimization problems.
\newblock {\em CoRR} abs/1412.6544.

\bibitem[\protect\citeauthoryear{He \bgroup et al\mbox.\egroup }{2015}]{He}
He, K.; Zhang, X.; Ren, S.; and Sun, J.
\newblock 2015.
\newblock Delving deep into rectifiers: Surpassing human-level performance on
  imagenet classification.
\newblock {\em CoRR} abs/1502.01852.

\bibitem[\protect\citeauthoryear{Keskar \bgroup et al\mbox.\egroup
  }{2016}]{DBLP:journals/corr/KeskarMNST16}
Keskar, N.~S.; Mudigere, D.; Nocedal, J.; Smelyanskiy, M.; and Tang, P. T.~P.
\newblock 2016.
\newblock On large-batch training for deep learning: Generalization gap and
  sharp minima.
\newblock {\em CoRR} abs/1609.04836.

\bibitem[\protect\citeauthoryear{Krizhevsky}{2009}]{CIFAR10}
Krizhevsky, A.
\newblock 2009.
\newblock Learning multiple layers of features from tiny images.
\newblock Technical report.

\bibitem[\protect\citeauthoryear{LeCun and
  Cortes}{}]{lecun-mnisthandwrittendigit-2010}
LeCun, Y., and Cortes, C.
\newblock Mnist handwritten digit database.

\bibitem[\protect\citeauthoryear{LeCun, Kavukcuoglu, and
  Farabet}{2010}]{lecun2010convolutional}
LeCun, Y.; Kavukcuoglu, K.; and Farabet, C.
\newblock 2010.
\newblock Convolutional networks and applications in vision.
\newblock In {\em Circuits and Systems (ISCAS), Proceedings of 2010 IEEE
  International Symposium on},  253--256.
\newblock IEEE.

\bibitem[\protect\citeauthoryear{{Li} \bgroup et al\mbox.\egroup
  }{2018}]{intrinsic}
{Li}, C.; {Farkhoor}, H.; {Liu}, R.; and {Yosinski}, J.
\newblock 2018.
\newblock {Measuring the Intrinsic Dimension of Objective Landscapes}.
\newblock {\em ArXiv e-prints}.

\bibitem[\protect\citeauthoryear{Rumelhart, Hinton, and
  Williams}{1986}]{fully_connected_original}
Rumelhart, D.~E.; Hinton, G.~E.; and Williams, R.~J.
\newblock 1986.
\newblock Parallel distributed processing: Explorations in the microstructure
  of cognition, vol. 1.
\newblock Cambridge, MA, USA: MIT Press.
\newblock chapter Learning Internal Representations by Error Propagation,
  318--362.

\bibitem[\protect\citeauthoryear{Spruill}{2007}]{MR2335894}
Spruill, M.~C.
\newblock 2007.
\newblock Asymptotic distribution of coordinates on high dimensional spheres.
\newblock {\em Electron. Comm. Probab.} 12:234--247.

\end{thebibliography}

\end{document}